\documentclass{IEEEtran}
\usepackage{cite}

\usepackage{authblk}
\usepackage{fancyhdr}

\usepackage{graphicx}
\graphicspath{ {figures/} }
\usepackage[skip=1pt]{caption}

\usepackage{booktabs}
\usepackage{mathtools}
\usepackage{multirow}
\usepackage{array}
\usepackage[table,xcdraw]{xcolor}

\usepackage{tabularx}
\usepackage{lipsum}
\usepackage{hyperref}
\usepackage[misc]{ifsym}

\usepackage[font=footnotesize]{caption}

\newcolumntype{P}[1]{>{\centering\arraybackslash}p{#1}}
\newcolumntype{?}{!{\vrule width 1pt}}

\fancypagestyle{firststyle}
{
   \fancyhf[L]{Kiourt, C., Kalles, Dimitris and Kanellopoulos, P., (2017), \textbf{How game complexity affects the playing behavior of synthetic agents}, 15th European Conference 
on Multi-Agent Systems, Evry, 14-15 December 2017}
  \lfoot{\textit{The final publication is available at Springer via \href{http://dx.doi.org/[insert DOI]}{http://dx.doi.org/[insert DOI]}} }
 \rfoot{- \thepage - }
}

\begin{document}
\raggedbottom 
\title{How game complexity affects the playing behavior of synthetic agents}
\author[1]{Chairi Kiourt}
\author[1]{Dimitris Kalles}
\author[2]{Panagiotis Kanellopoulos}

\affil[1]{School of Science and Technology, Hellenic Open University, Patras, Greece, chairik, kalles@eap.gr}
\affil[2]{CTI “Diophantus” and University of Patras, Rion, Greece.	kanellop@ceid.upatras.gr}

\maketitle
\thispagestyle{firststyle}
\begin{abstract}
Agent based simulation of social organizations, via the investigation of agents' training and learning tactics and strategies, has been inspired by the ability of humans to learn from social environments which are rich in agents, interactions and partial or hidden information. Such richness is a source of complexity that an effective learner has to be able to navigate. This paper focuses on the investigation of the impact of the environmental complexity on the game playing-and-learning behavior of synthetic agents. We demonstrate our approach using two independent turn-based zero-sum games as the basis of forming social events which are characterized both by competition and cooperation. The paper's key highlight is that as the complexity of a social environment changes, an effective player has to adapt its learning and playing profile to maintain a given performance profile.
\end{abstract}	

\section{Introduction}
Turn-based zero-sum games are most popular when it comes to studying social environments and multi-agent systems \cite{Ferber03fromagents, Shoham:2008, Wooldridge:2009:IMS:1695886}. For a game agent, the social environment is represented by a game with all its agents, components and entities, such as rules, pay-offs and penalties, amongst others \cite{Shoham:2008, Ferber:1999:MSI:520715, Kiourt:2016:AB}, while learning in a game is said to occur when an agent changes a strategy or a tactic in response to new information \cite{Kiourt:2016:AB, Marom:2001, AlKhateeb:2011, Caballero:2011:UCA:2027474.2027591}. Social simulation involves artificial agents with different characteristics (synthetic agents), which interact with other agents, possibly employing a mix of cooperative and competitive attitudes, towards the investigation of social learning phenomena\cite{Ferber:1999:MSI:520715, Kiourt:2016:AB, Gilbert:2005:SSS:1098701}.

The mimicking of human playing behaviors by synthetic agents is a realistic method for simulating game-play social events \cite{Kiourt:2016:AB}, where the social environment (games) as well as the other agents (opponents) \cite{Kiourt:2016:CIG, Kiourt:2016:EUMAS} are among the key factors which affect the playing behavior of the agents.

The solvability of board games is being investigated for over 25 years\cite{Allis:1988, Allis:1994searchingfor, Vandemherik:2002:277, Heule:2007:105}. Several studies focusing on board game complexity have shown that board games vary from low to high complexity levels \cite{Allis:1994searchingfor, Vandemherik:2002:277, Heule:2007:105}, which are mainly based on the game configuration and the state space of the game, with more complex games having larger rule set and more detailed game mechanics. In general, solvability is related to the \textit{state-space complexity} and \textit{game-tree complexity} of games\cite{Vandemherik:2002:277, Heule:2007:105}. The \textit{state-space complexity }is defined as the number of legal game positions obtainable from the initial position of the game. The \textit{game-tree complexity} is defined as the number of leaf nodes in the solution search tree of the initial position of the game. In our investigation, we adopted the \textit{state-space complexity} approach, which is the most-known and widely used \cite{Allis:1994searchingfor, Vandemherik:2002:277, Heule:2007:105}.

The complexity of a large set of well-known games has been calculated \cite{Vandemherik:2002:277, Heule:2007:105} at various levels, but their usability in multi-agent systems as regards the impact on the agents' learning/playing progress is still a flourishing research field.

In this article, we study the game complexity impact on the learning/training progress of synthetic agents, as well as on their playing behaviors, by adopting two different board games. Different playing behaviors \cite{Kiourt:2016:AB} are adopted for the agents' playing and learning progress. We experiment with varying complexity levels of\textit{ Connect-4 }(a medium complexity game) and \textit{RLGame} (an adaptable complexity game). These two different games cover an important range of diverse social environments, as we are able to experiment at multiple complexity levels, as determined by a legality-based model for calculating state-space complexity. Our experiments indicate that synthetic agents mimic quite well some human-like playing behaviors in board games. Additionally, we demonstrate that key learning parameters, such as exploitation-vs-exploration trade-off, learning backup and discount rates, and speed of learning are important elements for developing human-like playing behaviors for strategy board games. Furthermore, we highlight that, as the complexity of a social environment changes, the playing behavior (essentially, the learning parameters set-up) of a synthetic agent has to adapt to maintain a given performance profile.


\section{Background Knowledge}
In this section, we describe the games adopted for the experimental sessions, the structure of the synthetic agents' learning mechanisms and the development of the social environments.

\begin{figure}
\centering
\includegraphics[trim = 0 50 0 60, scale=0.15]{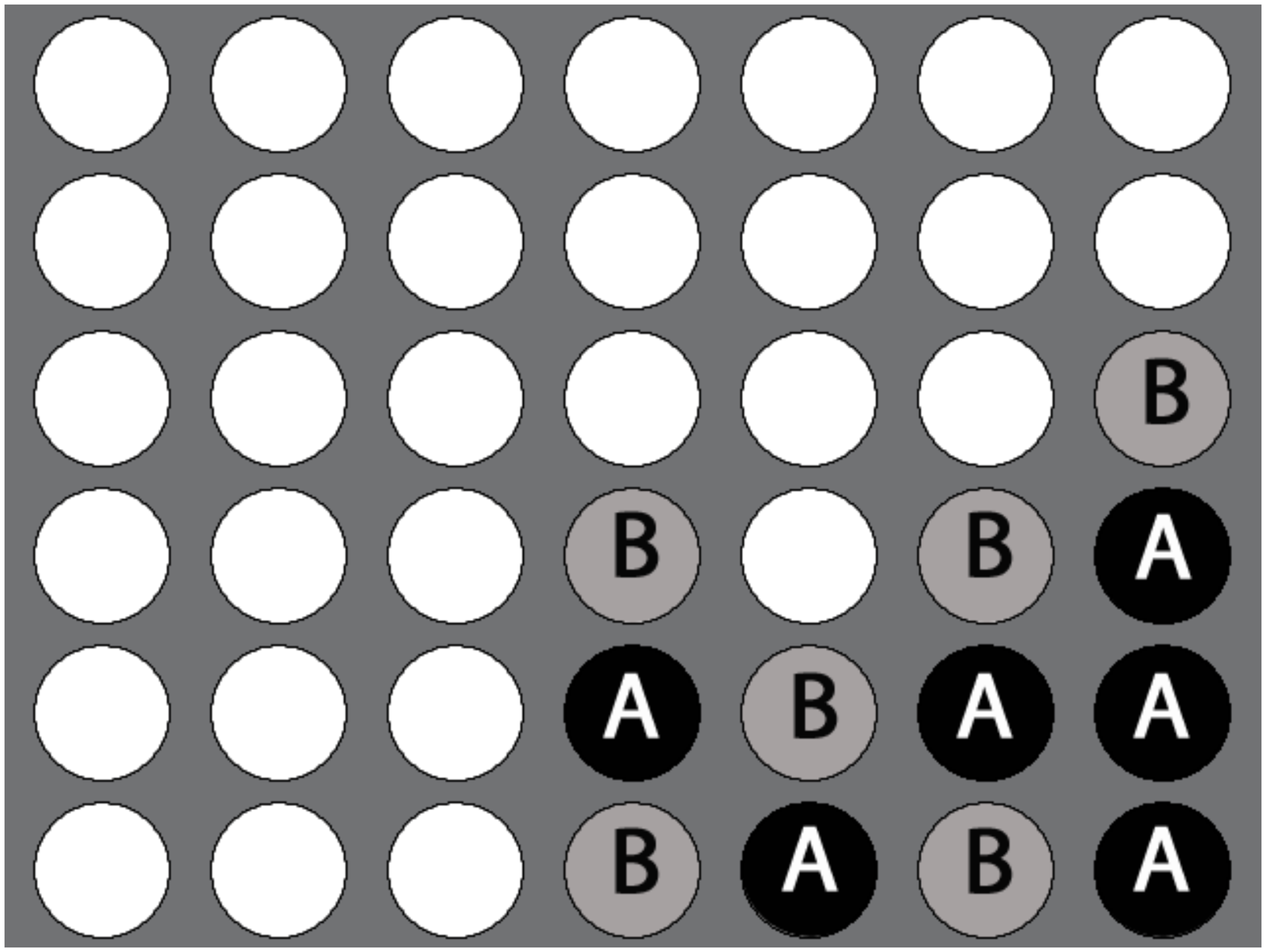}
\caption{A \textit{Connect-4} game in which player B wins.}
\label{fig:1}
\end{figure}

\textit{Connect-4} is a relatively recent game, fairly similar to \textit{tic-tac-toe}, but uses a  $6\times 7$ board with gravity. Both agents have 21 identical `coins', and each agent may only place its coins in the lowest available slot in a selected column (essentially, by inserting a coin at the free top of the column and allowing it to ``fall''). The goal of the game is to connect four of one's own coins of the same color next to each other vertically, horizontally or diagonally before the opponent reaches that goal. If all of both agents' coins are placed and no agent has achieved this goal, the game is a draw. \textit{Connect-4 }is a turn-based game and each agent has exactly one move per turn. It has a medium state space complexity of $4.5\times10^{12}$ board positions \cite{Edelkamp:2008}. Fig. \ref{fig:1} depicts an example of the \textit{Connect-4} game, in which agent \textit{B} wins the game.

\textit{RLGame} is a board game \cite{Kalles:2001:VGD:372202.372204} involving two agents and their pawns, played on a  square board. Two $\alpha \times\alpha$ square bases are on opposite board corners; these are initially populated by $\beta$ pawns for each agent, with the white agent starting from the lower left base and the black agent starting from the upper right one. The possible configurations of the \textit{RLGame} are presented in Table \ref{teble:1}. The goal for each agent is to move a pawn into the opponent's base or to force all opponent pawns out of the board (it is the player –not the pawn– who acts as an agent, in our scenario). The base is considered as a single square, therefore a pawn can move out of the base to any adjacent free square. Agents take turns and pawns move one at a time, with the white agent moving first. A pawn can move vertically or horizontally to an adjacent free square, provided that the maximum distance from its base is not decreased (so, backward moves are not allowed). A pawn that cannot move is lost (more than one pawn may be lost in one move). An agent also loses by running out of pawns.

\begin{table}
\centering
\caption{A description of game configurations}
\label{teble:1}
\setlength{\tabcolsep}{15pt}
\begin{tabular}{{c}{c}}
\toprule
Board size ($\boldsymbol{n}$) & 5, 6, 7, 8, 9, 10 \\
\hline
Base size ($\boldsymbol{\alpha}$) & 2, 3, 4 \\
\hline
Number of pawns ($\boldsymbol{\beta}$) & 1, 2, 3, 4, 5, 6, 7, 8, 9, 10 \\
\bottomrule
\end{tabular}
\end{table}

\begin{figure}
\centering
\includegraphics[trim = 0 180 0 190, width=9cm]{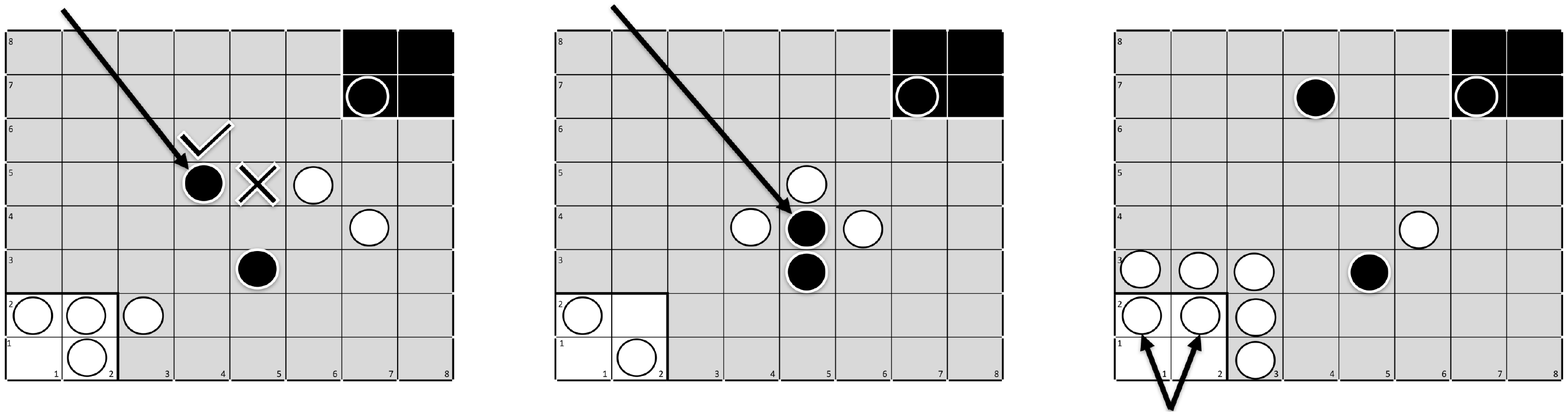}
\caption{Example of \textit{RLGame} rules into action.}
\label{fig:2}
\end{figure}

The implementation of some of the most important rules is depicted in Fig.\ref{fig:2}. In the leftmost board the pawn indicated by the arrow demonstrates a legal (``tick'') and an illegal (``cross'') move, the illegal move being due to the rule that does not allow decreasing the distance from the home (black) base. The rightmost boards demonstrate the loss of pawns, with arrows showing pawn casualties. A ``trapped'' pawn, either in the middle of the board or when there is no free square next to its base, automatically draws away from the game.

\begin{figure}
\centering
\includegraphics[trim = 0 100 0 100, width=6cm]{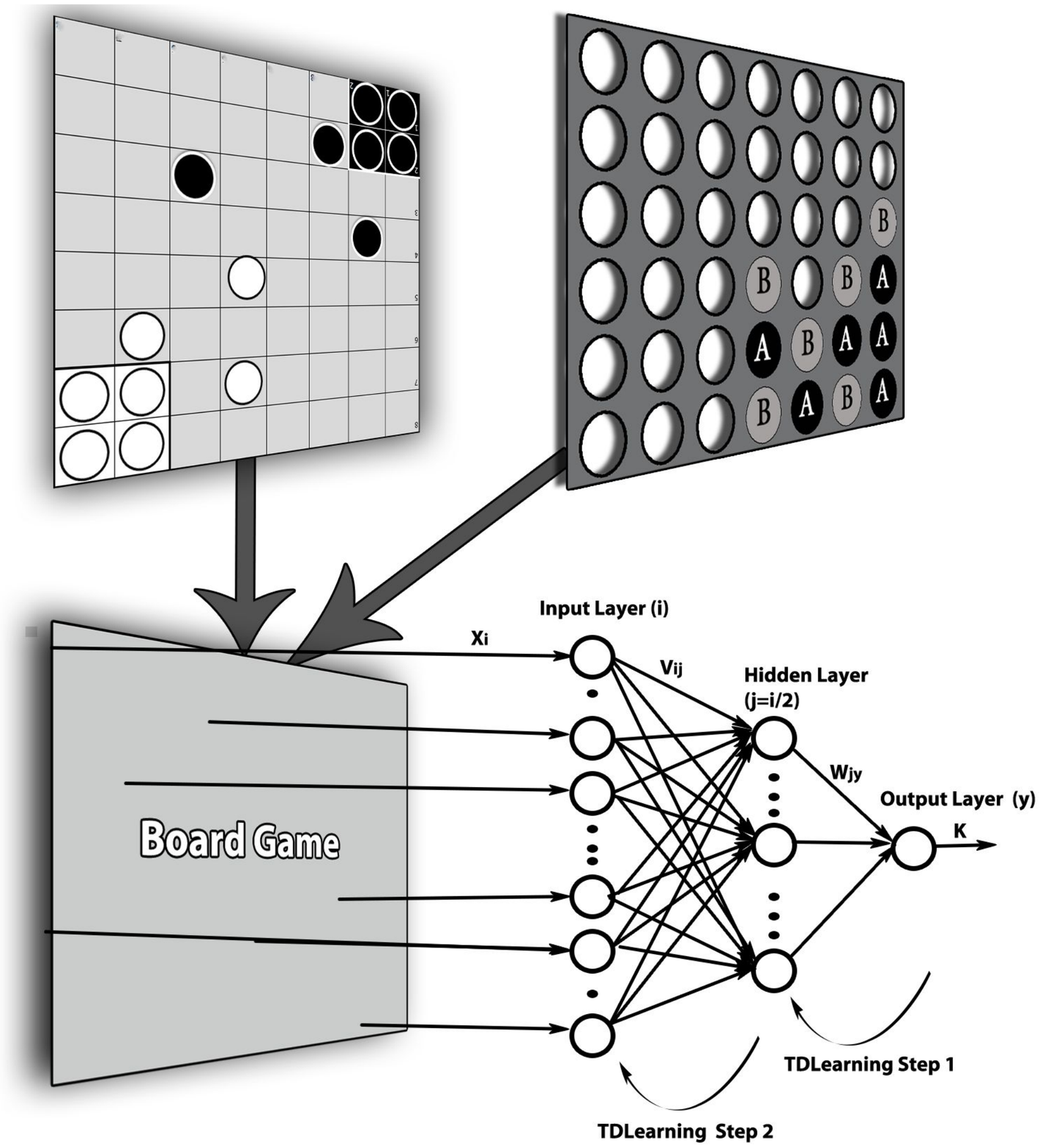}
\caption{Learning mechanism of \textit{RLGame} and \textit{Connect-4}}
\label{fig:3}
\end{figure}

For our study, in both games, each agent is an autonomous system that acts according to its characteristics and knowledge. The learning mechanism used (Fig. \ref{fig:3}) is based on reinforcement learning, by approximating the value function with a neural network \cite{Shoham:2008, Ferber:1999:MSI:520715}, as already documented in similar studies \cite{Tesauro1992,Tesauro:1995:TDL:203330.203343}.  Each autonomous (back propagation) neural network \cite{Sutton:1998:IRL:551283} is trained after each player makes a move. The board positions for the next possible move are used as input-layer nodes, along with flags regarding the overall board coverage. The hidden layer consists of half as many hidden nodes. A single node in the output layer denotes the extent of the expectation to win when one starts from a specific game-board configuration and then makes a specific move. After each move the values of the neural network are updated through the \textit{temporal difference learning method}, which is a combination of Monte Carlo and dynamic programming \cite{Sutton:1998:IRL:551283}. As a result, collective training is accomplished by putting an agent against other agents so that knowledge (experience) is accumulated.

For both games, the agent's goal is to learn an optimal strategy that will maximize the expected sum of rewards within a specific amount of time, determining which action should be taken next, given the current state of the environment. The strategy to select between moves is $\epsilon$-Greedy ($\epsilon$), with $\epsilon$ denoting the probability to select the best move (exploitation), according to present knowledge, and  $1-\epsilon$ denoting a random move (exploration) \cite{March:1991:EEO:2869931.2869936}. The learning mechanism is associated with two additional learning parameters, Gamma ($\gamma$) and Lambda ($\lambda$). A risky or a conservative agent behavior is determined by the γ parameter, which specifies the learning strategy of the agent and determines the values of future payoffs, with values in $[0,1]$ ; effectively, \textit{large values are associated with long-term strategies}. The speed and quality of agent learning is associated with $\lambda$, which is the learning rate of the neural network, also in $[0,1]$. \textit{Small values of $\lambda$ can result in slow, smooth learning; large values could lead to accelerated, unstable learning}. These properties are what we, henceforth, term as ``characteristic values'' for the playing agents.

\textit{RLGame }and \textit{Connect-4}, in their tournament versions \cite{Kiourt:2016:AB}, both fit the description of an autonomous organization and of a social environment, as defined by Ferber et al. \cite{Ferber03fromagents, Ferber:1999:MSI:520715}. Depending on the number of agents, social categories can be split into sub-categories of \textit{micro-social environments, environments composed of agent groups} and global societies, which are the next level of the cooperation and competition extremes of social organizations \cite{Shoham:2008, Ferber:1999:MSI:520715}.

On one hand, \textit{RLGame} was chosen because it is a fairly complex game for studying learning mechanism and developing new algorithms, because all the pawns of the game have the same playing attributes. It is not as complicated as \textit{Chess}, where almost all pieces have their own playing attributes, or \textit{Go}, which would make it difficult to study in detail the new learning algorithms. Furthermore, its complexity scales with the number of pawns and board dimensions, which allows for fewer non-linear phenomena that are endemic in games like \textit{Chess}, \textit{Go}, or \textit{Othello} (for example, knight movement in \textit{Chess} or column color inversion in \textit{Othello}, are both instances of such phenomena). We view this as a key facilitator in our quest for opponent modelling (but acknowledge the importance and the interestingness of non-linear aspects of game play). On the other hand,\textit{ Connect-4} was chosen due to its low complexity. With these two different games, we believe that we cover a quite important range of diverse environments, as we can accommodate several levels of complexity in \textit{RLGame} and pretty low complexity in \textit{Connect-4}.

\section{Game Complexity}
Combinatorial game theory provides several ways to measure the game complexity of two-person zero-sum games with perfect information \cite{Allis:1994searchingfor, Vandemherik:2002:277}, such as: \textit{state-space complexity, game tree size, decision complexity, game-tree complexity} and \textit{computational complexity}. In this study, we use the \textit{state-space complexity} approach, which is the most known and widely used \cite{Allis:1994searchingfor, Vandemherik:2002:277, Heule:2007:105}. Nowadays, dozens of games are solved by many different algorithms \cite{Vandemherik:2002:277, Heule:2007:105}.

\textit{Connect-4} is one of the first turn-based zero-sum games solved by computer \cite{Allis:1988}. It has a medium state space complexity of $4.5\times10^{12}$ board positions in $6\times7$ board size \cite{Edelkamp:2008}. Tromp \cite{Tromp:2008} presented some game theoretical values of \textit{Connect-4} on medium board sizes up to $width + height=15$, some of which are presented in Table \ref{table:2} \cite{Tromp:1995}.

\begin{table}[htb]
\setlength{\tabcolsep}{15pt}
\centering
\caption{\textit{Connect-4}, game configurations associated to their state-space sizes.}
\label{table:2}
\begin{tabular}{@{}clc@{}}
\toprule
\textbf{\begin{tabular}[c]{@{}c@{}}Height, Width\\ (size)\end{tabular}} & \textbf{\begin{tabular}[c]{@{}c@{}}State space\\ complexity\end{tabular}} & \textbf{\begin{tabular}[c]{@{}c@{}}$\boldsymbol{\beta}$\\ (coins per player)\end{tabular}} \\ \midrule
8,2                                                                     &                                      $\quad 1.33 \times 10^{4 }$                                   &                                    8                                        \\ \hline
8,3                                                                     &                                      $\quad 8.42 \times 10^{6 }$                                  &                                    12                                \\ \hline
8,4                                                                     &                                      $\quad 1.10 \times 10^9$                                   &                                     16                               \\ \hline
 7,4                                                                        &                                  $\quad 1.35 \times 10^8$                                   &                                    14                                     \\ \hline
 7,5                                                                      &                                    $\quad 1.42 \times 10^{10}$                                 &                                     17.5                                     \\ \hline
 6,5                                                                      &                                    $\quad 1.04 \times 10^9$                                   &                                    15                                       \\ \hline
 6,6                                                                    &                                      $\quad 6.92 \times 10^{10}$                                 &                                    18                                       \\ \hline
 6,7                                                                  &                                        $\quad 4.53 \times 10^{12}$                                  &                                    21                                       \\ \hline
 5,5                                                                       &                                   $\quad 6.98 \times 10^7$                                   &                                     12.5                                      \\ \hline
 5,6                                                                       &                                   $\quad 2.82 \times 10^9$                                   &                                     15                                      \\ \hline
  5,7                                                                   &                                      $\quad 1.13 \times 10^{10}$                                    &                                    17                                        \\ \bottomrule
\end{tabular}
\end{table}

The complexity of the \textit{RLGame} depends mainly on the value of parameters $n, \alpha,$ and $\beta$. The number of the various positions that might occur is bounded from above by:
\begin{equation}
\displaystyle\sum_{i=1}^{\beta} \sum_{j=1}^{\beta} \binom{n^2 - 2\alpha^2}{i+j}\binom{i+j}{i}(1+2(\beta-i))(1+2(\beta-j)).
\end{equation}

The first (leftmost) term denotes the number of ways to place $i+j$ pawns in the playing field (on the board but outside the bases) and the second term denotes the number of ways to partition these $i + j$  pawns into $i$ white and $j$ black pawns. The two rightmost terms intend to capture, for each given configuration of i white and j black pawns in the playing field, the additional number of positions that may occur because each player might have pawns in its own base and no pawn in the enemy base (there are $\beta - i$  such configurations for the white player) or a single pawn in the enemy base and possibly some pawns in its own base (again, there are $\beta - i$  such configurations for the white player).

Naturally, the above formula overestimates the number of possible states since it also includes illegal states, so we devised a simple simulation with the following steps to derive a better estimate.
\begin{itemize}
\item Given $n, \alpha,$ and $\beta$, we examine all valid configuration profiles $(n,\alpha,\beta,i,j)$ where $i,j$  denote the number of white and black pawns in the playing field.
\item We generated 1000 random positions per valid profile and tested whether some of them contained dead pawns (e.g., pawns with no legal moves).
\item For each configuration profile, we multiplied the fraction of such ``legit'' positions (we did not check whether a position without dead pawns can actually arise in a real game) with $[1 + 2 (\beta - i )][1 + 2 (\beta - j)]$  to take into account the $\beta - i$  (respectively, $\beta - j$ ) white (respectively, black) pawns that are not in the playing field for each configuration profile.
\item We summed the number of ``legit'' positions over all configuration profiles for the given values of $n, \alpha$ and $\beta$ and calculated the ratio of ``apparently legit states'' provided by this simulation over the ``theoretical estimation'' provided by the formula.
\end{itemize}

Table \ref{teble:1} reviews the $(n,\alpha,\beta)$  configurations we used; since bases should be at least one square apart in any given board, we eventually end-up with fewer valid $(n,\alpha)$  combinations (shown alongside the results in Table \ref{table:3}). Additionally, for valid configurations we demand that $0 < i \leq \beta$  and $0 < j \leq \beta$.

\begin{table}[htb]
\centering
\captionsetup{justification=centering}
\caption{\textit{RLGame}, games' extreme configurations associated to their state-space sizes. We state the theoretical upper bound and the ratio of ``legit'' positions that arose in simulations.}
\label{table:3}
\setlength{\tabcolsep}{8pt}
\begin{tabular}{@{}ccccc@{}}
\toprule§
\multirow{2}{*}{\textbf{Board, Base (size)}} & \multicolumn{2}{c}{\textbf{$\boldsymbol{\beta=1}$}}  & \multicolumn{2}{c}{\textbf{$\boldsymbol{\beta=10}$}} \\ \cmidrule(l){2-5}
                                             & \textbf{formula} & \textbf{ratio} & \textbf{formula} & \textbf{ratio} \\ \cmidrule(r){1-5}
5,2                                          & $3.83 \times  10^{2}$       & .991           & $1.11 \times  10^{10}$      & .127           \\ \hline
6,2                                          & $9.33 \times  10^{2}$       & .997           & $1.50 \times  10^{14}$      & .088           \\ \hline
7,2                                          & $1.89 \times  10^{3}$       & .999           & $6.93 \times  10^{17}$      & .177           \\ \hline
7,3                                          & $1.12 \times  10^{3}$       & .994            & $1.37 \times  10^{15}$      & .113           \\ \hline
8,2                                          & $3.43 \times  10^{3}$       & .998           & $7.21 \times  10^{20}$      & .373           \\ \hline
8,3                                          & $2.36 \times  10^{3}$       & 1.                & $9.10 \times  10^{18}$      & .254           \\ \hline
9,2                                          & $5.70 \times  10^{3}$       & .996           & $2.40 \times  10^{23}$      & .562           \\ \hline
9,3                                          & $4.29 \times  10^{3}$       & .997           & $9.64 \times  10^{21}$      & .486           \\ \hline
9,4                                          & $2.66 \times  10^{3}$       & 1.                & $3.72 \times  10^{19}$      & .315           \\ \hline
10,2                                         & $8.93 \times  10^{3}$       & 1.                & $3.50 \times  10^{25}$      & .712           \\ \hline
10,3                                         & $7.14 \times  10^{3}$       & 1.                 & $2.96 \times  10^{24}$      & .645           \\ \hline
10,4                                         & $4.97 \times  10^{3}$       & .998           & $5.12 \times  10^{22}$      & .530           \\ \bottomrule
\end{tabular}
\end{table}

 We only report the state-space size for the extreme cases of $\beta  = 1$ and $\beta  = 10$ for each $(n,\alpha)$  configuration used, since we observed that the approximation ratios strictly decrease with increasing values of $\beta$ (thus creating more room for pawn interdependencies which lead to illegal moves). These results are shown in Table \ref{table:3} and confirm that, even for relatively small dimensions, state space complexity is well over $10^{10}$.

\section{Experimental Sessions}
In order to study the game complexity effect in synthetic agents' learning/training process as well as in their playing behaviors, in multi-agent social environments, three independent tournament sessions (experiments) with the same pre-configurations were designed and run for both \textit{RLGame }and \textit{Connect-4}; for simplicity, we will name these tournament sessions as $RL-R(x\times y)$  for \textit{RLGame} and $C4-R(x\times y)$  for \textit{Connect-4}, where $(x\times y)$ presents the game configuration. Table \ref{table:4} presents the game configurations selected for the tournament sessions (experiments). We chose three different game configurations for each game, in order to study three different complexity level of each game. We remark that in the following we only compare agents playing the same game; we never compare an agent from \textit{RLGame} to an agent from \textit{Connect-4}.


\begin{table*}[htb]
\centering
\caption{Selected game configurations for the tournament sessions (experiments).}
\label{table:4}
\begin{tabular}{@{}P{2cm} P{2.2cm} P{1.8cm}| P{2cm} P{1.9cm} P{2cm}@{}}
\multicolumn{3}{c}{\textbf{\textit{Connect-4}}}                                                & \multicolumn{3}{c}{\textit{\textbf{RLGame}}}                                                        \\ \toprule
\scriptsize{\textbf{Experiment (Tournament) name}} & \scriptsize{\textbf{Size \quad\quad\tiny{(Height, Width)}}} & \scriptsize{\textbf{State space complexity}} & \scriptsize{\textbf{Experiment (Tournament) name}} & \scriptsize{\textbf{Size \quad\quad\tiny{(Board, Base)}}} & \scriptsize{\textbf{State space complexity ($\boldsymbol{\beta=10}$)}} \\ \toprule
$C4-R(8\times3$)                    & 8,3                  & $8.42 \times 10^6$            & $RL-R(5\times2)$                    & 5,2                & $1.11 \times 10^{10} $                  \\ \hline
$C4-R(7\times4)$                    & 7,4                  & $1.35 \times 10^8$             & $RL-R(7\times2)$                    & 7,2                & $6.93 \times 10^{17} $                  \\ \hline
$C4-R(6\times7)$                    & 6,7                  & $4.53 \times 10^{12}$            & $RL-R(10\times2)$                   & 10,2               & $3.50 \times 10^{25} $                  \\ \bottomrule
\end{tabular}
\end{table*}

According to the scenario of these tournaments sessions, we initiated 64 agents in a\textit{ Round Robin }tournament with 10 games per match. All agents had different characteristic value configurations for $\epsilon, \gamma$ and $\lambda$, with values ranging from 0.6 to 0.9, with an increment step of 0.1. Four different values for each characteristic value ($\epsilon$-$\gamma$-$\lambda$), implies $4^3  = 64$  agents with different playing behaviors (different characteristic values). Each agent played 63 matches against different agents, resulting in a total number of $\binom{64}{1} \times 10 = 200,160$  games, for each tournament session. All tournament sessions were identical in terms of agent configurations and flow of execution.

The ranges of the characteristic values ($\epsilon$-$\gamma$-$\lambda$) are selected, because of their association with the playing behaviors of the agent \cite{Kiourt:2016:AB}. For example, if we had an agent that exploits 5\% of its knowledge ($\epsilon$), then it almost always learn something new and would only rarely demonstrate what it learned \cite{Sutton1988,Sutton:1998:IRL:551283}. Also, if we set $\lambda = 0.05$, the agent would learn very slow, which is not effective in case the opponent opts to play head-on attack (one pawn moving directly to the opponent base for \textit{RLGame}), as an agent with a low $\lambda$ may be less interested to learn a more structured strategy by using many pawns that may defend its base or to force opponent pawns out of the board. Wiering et al. \cite{Wiering2005LearningTP} suggested that $\lambda$ values larger than 0.6 perform best. The discount rate parameter, $\gamma$, as reported by Sutton and Barton \cite{Sutton:1998:IRL:551283}, tilts the agent towards being myopic and only concerned with maximizing immediate rewards when $\gamma = 0$, while it allows the agent to become more farsighted and take future rewards into account more strongly when $\gamma = 1$. For this reason, on one hand, by setting the $\gamma$ values roughly to 0.6, we may say that the agent adopts short term strategies (risky), on the other hand, by setting the $\gamma$ values to 0.9 we represent the agents with long term strategies (conservative agents). With the characteristic values $\epsilon$-$\gamma$-$\lambda$ ranging between 0.6 and 0.9, we kept a balance.

Based on the agents' characteristic values ($\epsilon$-$\gamma$-$\lambda$) and their performance, we developed a set of playing behavior descriptors \cite{Kiourt:2016:AB}, see Table \ref{table:5}.


\begin{table}[htb]
\centering
\captionsetup{justification=centering}
\caption{Agents playing behavior descriptors based on their characteristic values and their performance}

\label{table:5}
\begin{tabular}{@{}cl@{}}
\toprule
\textbf{Characteristic}                                           & \textbf{Key parameters}                                           \\
\textbf{Values}                                                   & \textbf{(Playing behavior descriptors)}                           \\ \midrule
\rowcolor[HTML]{C0C0C0}
\cellcolor[HTML]{C0C0C0}{\color[HTML]{000000} }                   & {\color[HTML]{000000} \textit{Exploration, exploitation tradeoff}}         \\
\rowcolor[HTML]{C0C0C0}
\multirow{-2}{*$\boldsymbol{0.6 \leq \epsilon\leq0.9}$}{\cellcolor[HTML]{C0C0C0}{\color[HTML]{000000} }} & {\color[HTML]{000000} \textbf{(\textit{knowledge explorer to exploiter})}} \\
                                                                  & \textit{Learning back-up and discount rates}                               \\
\multirow{-2}{*$\boldsymbol{0.6 \leq \gamma\leq0.9}$}{}                                                & \textbf{(\textit{risky to conservative, short to long term strategies})}   \\
\rowcolor[HTML]{C0C0C0}
\cellcolor[HTML]{C0C0C0}                                          & \textit{Speed \& stability of learning }                                   \\
\rowcolor[HTML]{C0C0C0}
\multirow{-2}{*$\boldsymbol{0.6 \leq \lambda\leq0.9}$}{\cellcolor[HTML]{C0C0C0}}                        & \textbf{(\textit{slow smooth to fast and unstable learning})}              \\
                                                                  & \textit{Agents' rankings, performance}                                     \\
\multirow{-2}{*$\boldsymbol{1 \leq r \leq64}$}{}                                                & \textbf{(\textit{good playing to bad playing agents})}                     \\ \cmidrule(l){1-2}
\end{tabular}
\end{table}

The first three descriptors are composed from the characteristic values derived from previous experiments \cite{Kiourt:2016:AB}. Those three descriptors define the characteristics limits, which determine playing behaviors depending on their preferred strategies. Simply put, every descriptor may represent a synthetic agent's playing behavior in the experimental social environment. An example of synthetic agent's playing behavior is that a `Knowledge Exploiter' (high $\epsilon$ value) and `Conservative' (high $\gamma$ value) and `Fast, Unstable Learner' (high $\lambda$ values) agent tends to be `Bad playing' (high \textit{r} value), which we do not consider positive for a game-playing agent.

The agents are rated by using the \textit{ReSkill} tool \cite{Kiourt:2016:ReSkill}. All the last ratings of tournament sessions are converted to rankings ($r$), in order to compare more effectively the experiments by using statistical methods, such as the \textit{Spearman's rank correlation coefficient} ($\rho$) \cite{Spearman:1904}, which measures the statistical dependence between two variables, and is specifically efficient at capturing the monotonic (non-linear, in general) correlation on ranks and the \textit{Kendall rank correlation coefficient} ($\tau$) \cite{Kendall:1936}, which measures the ordinal association between two measured quantities, both considered as adequate statistical measures to compare ranking lists quantitatively \cite{Langville:2012}. As known, the range of both coefficients falls within $[-1,1]$, with high negative values representing strong negative correlation, low absolute values representing small or no correlation and high positive values representing strong positive correlation. Table \ref{table:6} shows a \textit{Spearman's }and \textit{Kendall's} correlation coefficients distance heat-map, for the tournament sessions introduced in Table \ref{table:4}. The top value of each cell shows the $\rho$ correlation coefficient while the bottom value of each cell the τ correlation coefficient. Darker gray cells indicate a high correlation between two tournament sessions (agent rankings), while lighter gray cells indicate a strong negative correlation. Table \ref{table:6} also represents an indicative correlation between the state-space complexities of the social environments.

\begin{table*}[htb]
\centering
\captionsetup{justification=centering}
\caption{\textit{Spearman's} and \textit{Kendall's} correlation coefficients comparison of each tournament session, presented as a distance heat-map, where high distances are presented with light gray and smaller distances with darker gray}
\label{table:6}
\resizebox{\textwidth}{!}{
\begin{tabular}{c?c|c|c|c|c|c}
\hline
                     & \textbf{$\boldsymbol{C4-R(8\times3)}$}                          & \textbf{$\boldsymbol{C4-R(7\times4}$)}                          & \textbf{$\boldsymbol{C4-R(6\times7}$)}                          & \textbf{$\boldsymbol{RL-R(5\times2}$)}                          & \textbf{$\boldsymbol{RL-R(7\times2}$)}                          & \textbf{$\boldsymbol{RL-R(10\times2}$)}                         \\
                     & \textbf{$\boldsymbol{8.42 \times 10^6}$}                         & \textbf{$\boldsymbol{1.35 \times 10^8}$}                         & \textbf{$\boldsymbol{4.53 \times 10^{12}}$}                        & \textbf{$\boldsymbol{1.11 \times 10^{10}}$}                        & \textbf{$\boldsymbol{6.93 \times 10^{17}}$}                        & \textbf{$\boldsymbol{3.50 \times 10^{25}}$}                        \\ \toprule
\textbf{$\boldsymbol{C4-R(8\times3})$}   & \cellcolor[HTML]{656565}                    & \cellcolor[HTML]{C0C0C0}0.340               & \cellcolor[HTML]{EFEFEF}-0.043              & \cellcolor[HTML]{EFEFEF}-0.340              & \cellcolor[HTML]{EFEFEF}-0.274              & \cellcolor[HTML]{EFEFEF}-0.192              \\
\textbf{$\boldsymbol{8.42 \times 10^6}$}  & \multirow{-2}{*}{\cellcolor[HTML]{656565}1} & \cellcolor[HTML]{C0C0C0}0.237               & \cellcolor[HTML]{EFEFEF}-0.019              & \cellcolor[HTML]{EFEFEF}-0.212              & \cellcolor[HTML]{EFEFEF}-0.186              & \cellcolor[HTML]{EFEFEF}-0.134              \\ \hline
\textbf{$\boldsymbol{C4-R(7\times4)}$}   & \cellcolor[HTML]{C0C0C0}0.340               & \cellcolor[HTML]{656565}                    & \cellcolor[HTML]{EFEFEF}0.090               & -0.477                                      & -0,5                                        & -0.518                                      \\
\textbf{$\boldsymbol{1.35 \times 10^8}$}  & \cellcolor[HTML]{C0C0C0}0.237               & \multirow{-2}{*}{\cellcolor[HTML]{656565}1} & \cellcolor[HTML]{EFEFEF}0.064               & -0.332                                      & 0.362                                       & -0.362                                      \\ \hline
\textbf{$\boldsymbol{C4-R(6\times7)}$}   & \cellcolor[HTML]{EFEFEF}-0.043              & \cellcolor[HTML]{EFEFEF}0.090               & \cellcolor[HTML]{656565}                    & \cellcolor[HTML]{EFEFEF}-0.179              & \cellcolor[HTML]{EFEFEF}-0.167              & \cellcolor[HTML]{EFEFEF}-0.229              \\
\textbf{$4\boldsymbol{.53 \times 10^{12}}$} & \cellcolor[HTML]{EFEFEF}-0.019              & \cellcolor[HTML]{EFEFEF}0.064               & \multirow{-2}{*}{\cellcolor[HTML]{656565}1} & \cellcolor[HTML]{EFEFEF}-0.127              & \cellcolor[HTML]{EFEFEF}-0.121              & \cellcolor[HTML]{EFEFEF}-0.138              \\ \hline
\textbf{$\boldsymbol{RL-R(5\times2)}$}   & \cellcolor[HTML]{EFEFEF}-0.340              & -0.477                                      & \cellcolor[HTML]{EFEFEF}-0.179              & \cellcolor[HTML]{656565}                    & \cellcolor[HTML]{9B9B9B}0.673               & \cellcolor[HTML]{9B9B9B}0.720               \\
\textbf{$\boldsymbol{1.11 \times 10^{10}}$} & \cellcolor[HTML]{EFEFEF}-0.212              & -0.332                                      & \cellcolor[HTML]{EFEFEF}-0.127              & \multirow{-2}{*}{\cellcolor[HTML]{656565}1} & \cellcolor[HTML]{9B9B9B}0.482               & \cellcolor[HTML]{9B9B9B}0.519               \\ \hline
\textbf{$\boldsymbol{RL-R(7\times2)}$}   & \cellcolor[HTML]{EFEFEF}-0.274              & -0.500                                      & \cellcolor[HTML]{EFEFEF}-0.167              & \cellcolor[HTML]{9B9B9B}0.673               & \cellcolor[HTML]{656565}                    & \cellcolor[HTML]{9B9B9B}0.740               \\
$\boldsymbol{6.93 \times 10^{17}}$          & \cellcolor[HTML]{EFEFEF}-0.186              & -0.362                                      & \cellcolor[HTML]{EFEFEF}-0.121              & \cellcolor[HTML]{9B9B9B}0.482               & \multirow{-2}{*}{\cellcolor[HTML]{656565}1} & \cellcolor[HTML]{9B9B9B}0.561               \\ \hline
\textbf{$\boldsymbol{RL-R(10\times2)}$}  & \cellcolor[HTML]{EFEFEF}-0.192              & -0.518                                      & \cellcolor[HTML]{EFEFEF}-0.229              & \cellcolor[HTML]{9B9B9B}0.720               & \cellcolor[HTML]{9B9B9B}0.740               & \cellcolor[HTML]{656565}                    \\
\textbf{$\boldsymbol{3.50 \times 10^{25}}$} & \cellcolor[HTML]{EFEFEF}-0.134              & -0.362                                      & \cellcolor[HTML]{EFEFEF}-0.138              & \cellcolor[HTML]{9B9B9B}0.519               & \cellcolor[HTML]{9B9B9B}0.561               & \multirow{-2}{*}{\cellcolor[HTML]{656565}1}\\ \bottomrule
\end{tabular}}
\end{table*}

\begin{figure*}
\centering
\includegraphics[trim = 80 0 80 0, height=18cm, angle =-90]{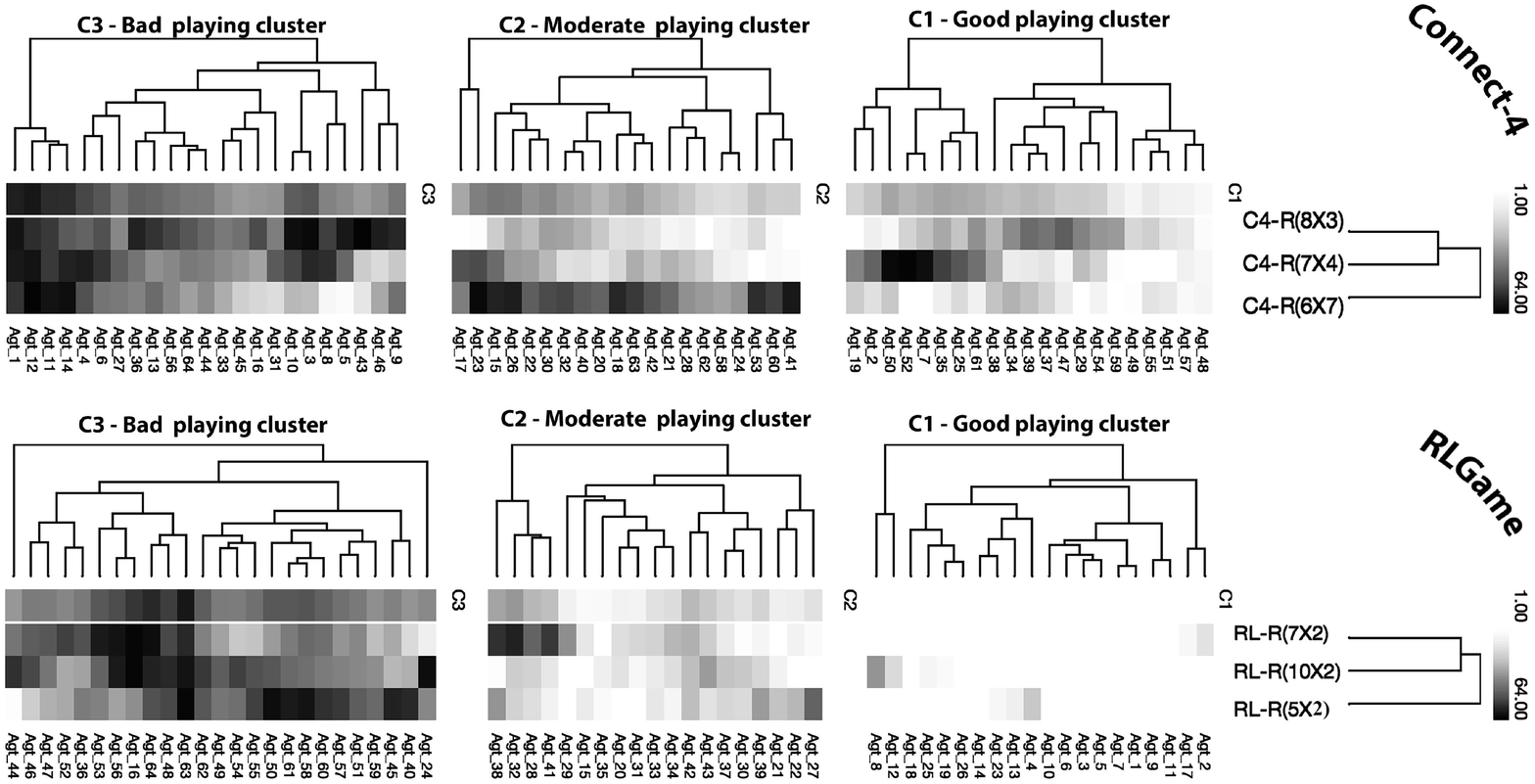}
\captionsetup{justification=centering}
\caption{\textit{K-means} clustering for each tournament session of both games and a dendrogram representing the correlation of agents and tournament sessions}
\label{fig:4}
\end{figure*}

In order to verify the tournament sessions' correlations, we applied a \textit{k-means} clustering for all tournament sessions and we developed the heat-maps of Fig. \ref{fig:4}. We set the number of the \textit{k-means} clusters fixed to $3$ (C1, C2 and C3), to build three clusters based on the agents' performance (by using the agents' rankings from each tournament sessions). Also, we set the re-runs of the \textit{k-means} algorithm to 100 and the \textit{maximal iterations} within each algorithm run to $300$. Due to the number of the agents ($64$) and the number of the tournament sessions ($3$ for each game), the \textit{k-means} configuration was good  enough to show the best correlation between the agents' performances associated to the tournament sessions. We tested the\textit{ k-means} algorithm with larger number of \textit{re-runs }and \textit{maximal iteration} but there was no difference in the result. Fig. \ref{fig:4} presents three rows for each game (one for each tournament session) and 64 columns (one for each agent). The columns are separated in three clusters for each game. Each cluster (C1, C2 and C3) depicts the association of the agents, based on their rankings in the three tournament sessions. Each agent (rows in the graphs) is composed from three colored cells, where each cell depicts the performance of the agent in the corresponding tournament session. The colored bars, from light grey to dark grey, at the right of each graph, depict the ranking positions. In example, each dark gray cell depicts a bad playing agent in the corresponding tournament (row), the darkest cell of the C3 cluster, tournament session $C4-R(8\times3)$, shows the worst playing agent of that experimental session, which was ranked in the $64$\textsuperscript{th} position in the last round of the tournament. The correlation between the agents of each cluster (C1, C2 and C3), of each game, is depicted by a tree graph (dendrogram) in the top of each cluster. Each row (tournament session) and column (agents) are clustered by leaf ordering. As leaves we mean the lines (leaf of the dendrogram) that show the correlation between two variables (agents or tournament sessions). For example, the leaves: $C4-R(8\times3)$ and $C4-R(7\times4)$  are higher related (rows of \textit{Connect-4} game), than the leaf $C4-R(6\times7)$, which differs more than the two other leaves. This can be confirmed if one checks the color shades of the cells (agent) in the three tournaments (three cells in a row). If an agent has similar color shades in the three cells, it means that the agent performs the same in the three tournament sessions of the game. For example, the top performer agent of C1 cluster in \textit{Connect-4} game is \textit{Agt\_48}, each cell of each tournament session has intense light gray color.

\begin{figure*}
\centering
\includegraphics[trim = 0 100 0 100, width=18 cm]{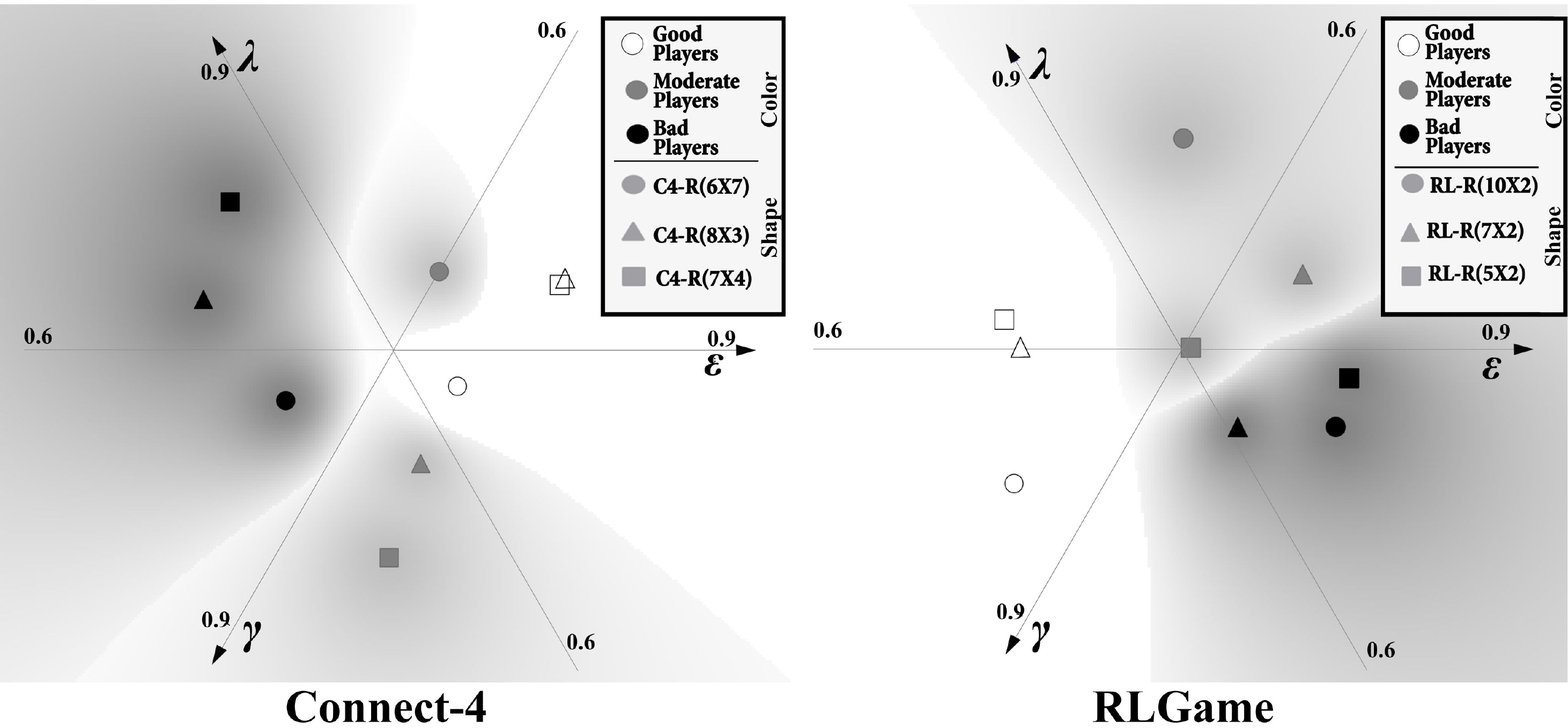}
\captionsetup{justification=centering}
\caption{Spatial allocation of the cluster (C1, C2 and C3), associated to the characteristic values ($\epsilon$-$\gamma$-$\lambda$), of both games (\textit{Connect-4} and \textit{RLGame}).}
\label{fig:5}
\end{figure*}

Fig. \ref{fig:5} depicts the spatial allocation of each cluster, resulting from the \textit{k-means} clustering (Fig. \ref{fig:4}), associated to the average number of the agents' characteristic values ($\epsilon$-$\gamma$-$\lambda$), respectively for each game (\textit{Connect-4} and \textit{RLGame}). The shapes in the graphs in Fig. \ref{fig:5} indicate the state-space complexity of the different tournament sessions of each game. The circles represent the high state-space complexities, triangles represent the medium state-space complexities and the squares represent the low state-space complexities respectively for each game. The colors of the shapes represent the C1, C2 and C3 clusters. In example, the black square in the left graph depicts the C3 cluster (bad playing agents) of the \textit{Connect-4's} lowest state space complexity, in a special allocation of the characteristic values ($\epsilon$-$\gamma$-$\lambda$). This means that the bad playing agents of the \textit{Connect-4's} lowest complexity, seem to have low $\epsilon$-greedy $(\epsilon \approx 0.68)$, high lambda $(\lambda \approx 0.85)$  and medium gamma $(\gamma \approx 0.72)$. If we associate these characteristic values with the playing behaviour descriptors of Table \ref{table:5}, we can say that a bad playing agent in a low complexity environment of the \textit{Connect-4} game, seems to be an ``exploiter'', a ``fast, unstable learner'', which takes into account ``medium-term strategies''.

\section{Discussion}
The correlation coefficient analysis that compared all the tournament sessions of both games (Table \ref{table:6}) shows a high correlation coefficient between the three tournament sessions of \textit{RLGame}. The correlation coefficient between two experiments (two different tournament sessions) presents the similarity or the differentiation of the agents' performances (agents with the same playing profile) in the studied experimental state spaces. \textit{Connect-4's} tournament sessions show a quite good correlation between the two lower complexity state-spaces ($C4-R(8\times3)$  and $C4-R(7\times4)$ ), while the correlation of the higher complexity state space compared to the two lower complexity state-spaces of the Connect-4 appears to be neutral, with about 0 correlation coefficient ($C4-R(8\times3$) and $C4-R(7\times4)$  correlation compared to $C4-R(6\times7)$. An important highlight is, that while the complexity of the \textit{Connect-4} increases, the negative correlation between the \textit{Connect-4's} and \textit{RLGame's} tournament sessions decreases (third column and last three rows of Table 6). For example, the correlation between $C4-R(8\times3)$  and all \textit{RLGame} tournament sessions show an average $\rho \approx -0.268$  and $\tau \approx -0.177$, while the correlation between the $C4-R(6\times7)$  and all \textit{RLGame} tournament sessions, shows an average $\rho \approx -0.191$  and $\tau \approx -0.128$, which is an increase of 4\% for ρ and 3\% for τ correlations. This highlights that as the complexity level of \textit{Connect-4} increases (referring to the $C4-R(6\times7)$ variant), stronger positive correlation with all the tournament session of \textit{RLGame} is observed, as both $\rho$ and $\tau$ values increase from negative to $0$. Generally in \textit{RLGame}, agents with similar playing profiles behave in the same way as the state complexity of \textit{RLGame} changes, while this is not the case for agents in \textit{Connect-4}. We had originally reported that we attributed the differences in performance of agents of the same set-up to the different complexity of the \textit{Connect-4} and \textit{RLGame} games. This is further strengthened by the finding that a \textit{Connect-4} variant of higher complexity is closer to \textit{RLGame}.

The \textit{k-means }clustering shows a higher correlation between the \textit{RLGame} tournament sessions than the corresponding \textit{Connect-4's} tournament sessions, which is depicted by the heat-maps of Fig. \ref{fig:4} and supports the results of correlation coefficient analysis. The color shades (heat-maps) of the \textit{RLGame} tournament sessions are more evenly allocated compared to the heat-maps of the \textit{Connect-4} tournament sessions. The single most uneven color allocation of the \textit{Connect-4's} heat-maps appears in the C2 cluster, where one mostly finds moderate playing agents and highlights that almost all agents of this cluster played better in the two lower levels the\textit{ Connect-4 }state-space complexity variants.

The special allocations of the clusters C1, C2 and C3 (for both games), associated with the characteristic values ($\epsilon$-$\gamma$-$\lambda$) and the performance of the cluster, highlight an estimation of the synthetic agents' playing behaviors of each cluster, as shown in Fig. \ref{fig:5}. For example, the good playing agents of the two lowest state-space complexities configurations of \textit{Connect-4} game (C1 clusters of $C4-R(8\times3)$  and $C4-R(7\times4)$ ), tend to have high $\epsilon$-greedy ($\epsilon \approx 0.81 \implies $\textit{knowledge exploiters}), medium lambda ($\lambda \approx 0.76 \implies $ \textit{medium speed  learner}) and small gamma ($\gamma \approx 0.69 \implies$ \textit{risky (short term strategy selection)}). The two graphs of Fig. \ref{fig:5} highlight important differences in the agents' performance and playing behaviors based on the games and their complexity variations, such as:

\begin{itemize}
\item Good playing agents tend to be exploiters (high $\epsilon$ value) in \textit{Connect-4}, in contrast to \textit{RLGame}, where good playing agents tend to be explorers (low $\epsilon$ value), which is reasonable since \textit{RLGame} is much more complex than \textit{Connect-4} and the good playing agents respond to the environment, thus shifting towards becoming knowledge explorers.
\item Bad playing agents are associated with low $\epsilon$ values in \textit{Connect-4} and high $\epsilon$ values in \textit{RLGame}, which is exactly the opposite to the good playing agents in both games.
\item Moderate playing agents are scattered in both graphs (both games) and their playing behaviors is not clear.
\end{itemize}

It is clear that the performance of the agents depends on the game and on its complexity level. Due to the higher complexity level of the \textit{RLGame}, the good playing agents need to be more sophisticated (more knowledge explorers, slow and smooth learners and focusing on longer term strategies), which is not surprising if one aims at a more realistic simulation of playing behavior.

Each good playing agents' cluster changes its characteristic values ($\epsilon$-$\gamma$-$\lambda$), only by slight shifting (as in Fig. \ref{fig:5}), as the complexity of the game increases. By observing the C1 clusters of the two lower complexity tournament sessions of both games $(C4-R(8\times3)$  and $C4-R(7\times4)$  for \textit{Connect-4}, $RL-R(5\times2)$  and $RL-R(7\times$2)  for \textit{RLGame}), we highlight that they have similar playing characteristic ($\epsilon$-$\gamma$-$\lambda$) values (the white triangles and squares are allocated to almost the same part, respectively, of each graph in Fig. \ref{fig:5}). The C1 cluster (white circle in left graph of Fig. \ref{fig:5}) of \textit{Connect-4's} highest complexity tournament session ($C4-R(6\times7)$) shows a slight shifting in comparison to the C1 clusters of the lower complexity tournament sessions $(C4-R(8\times3)$  and $C4-R(7\times4))$. We observe a shifting of about -12\% for $\epsilon$, -2\% for $\lambda$ and +8\% for $\gamma$.

The C1 cluster (white circle in right graph of Fig. \ref{fig:5}) of the \textit{RLGame's} highest complexity tournament session $(RL-R(10\times2))$, shows a similar slight sifting, in comparison to the C1 clusters of the lower complexity tournament sessions $(RL-R(5\times2)$  and $RL-R(7\times2))$. A shifting of about -2\% for $\epsilon$, -7\% for $\lambda$ and +3\% for $\gamma$ is observed.

Such shifting of the $\epsilon, \gamma$ and $\lambda$ values indicates that as the complexity of the environment increases (environments of \textit{Connect-4} and \textit{RLGame}), good playing agents tend to become more sophisticated (\textit{more knowledge explorers, more slow and smoother learners and focused in longer-term strategies}). The largest shifting appears in the C2 clusters (moderate playing agents) of both games' all complexity levels, which indicates that the moderate playing agents are hard to classify based on their characteristic values ($\epsilon$-$\gamma$-$\lambda$). The C3 clusters of the \textit{Connect-4} seem to be more affected by low $\epsilon$ values, while the C3 clusters of the \textit{RLGame} seem to be more affected by high $\epsilon$ values.

\section{Conclusion and Future Directions}
Based on the outcomes of the experimental tournament sessions, which spanned three different complexity levels for each game, \textit{Connect-4} and \textit{RLGame}, where we used the same agents' playing profile setups (same characteristic values $\epsilon$-$\gamma$-$\lambda$), we highlighted that an agents' playing profile does not readily lead to a comparable performance when the complexity of the environment (game) changes.

If an agent focuses on a specific performance level, in environments of varying complexity, its playing profile (characteristic values $\epsilon$-$\gamma$-$\lambda$) has to be re-adapted along specific directions based on the environment complexity. Our findings suggest that, as complexity increases (from \textit{Connect-4} to \textit{RLGame} and from a low-complexity \textit{RLGame} variant to a higher complexity one), an agent stands a better chance of maintaining its performance profile (as indicated by its ranking), by decreasing its $\epsilon$ and $\lambda$ values and increasing its $\gamma$ one (though, of course, the exact change ratios may be too elusive to define). For this reason, we state that the re-adaptation of the agents' characteristic values depends on the game and its complexity but, broadly speaking, we note that as the complexity of the environment increases, good playing agents have to be more sophisticated: increasing their knowledge exploration bias (lower $\epsilon$ values), becoming slower and smoother learners (lower $\lambda$ values) and focusing on longer term strategies (higher $\gamma$ values). These findings are corroborated by the experimental sessions of both games, \textit{Connect-4 }and \textit{RLGame} and it appears that an agent with a given $\epsilon$-$\gamma$-$\lambda$ profile cannot expect to maintain its performance profile if the environment changes with respect to the underlying complexity. Experimenting with a \textit{Connect-4} variant of large $n \times m$  dimensions and maybe extending \textit{Connect-4} to \textit{Connect-k} could eventually shift the association with \textit{RLGame} to larger positive values, thus further strengthening the validity of our findings.

The experimental results of this paper highlight that synthetic agents are important elements of the simulation of realistic social environments and that just a handful of characteristic values ($\epsilon$-$\gamma$-$\lambda$), namely, the exploitation-vs-exploration trade-off, learning backup and discount rates, and speed of learning, can synthesize a diverse population of agents with starkly different learning and playing behaviors.

An apparently promising and interesting investigation direction concerns the synthetic agents' application to other games (better known ones) and other complexity levels, such as checkers, chess etc., to investigate the learning progress of the synthetic agents' and the adjustability of their playing behaviors in diverse social environments. Additionally, as we highlighted that a synthetic agent's playing behavior may have to change in response to a change in the environment's complexity, this raises the generic question of how to modify one's characteristic values ($\epsilon$-$\gamma$-$\lambda$) based on an assessment of the surrounding environment. Such an assessment could be based either on the complexity of the environment or on the level of the opponent but both approaches involve making an estimation based on limited information (for example, a limited number of games against some opponents should be able to help an agent to gauge whether it operates in a complex or simple environment or where its opponents might be situated in terms of their values in the $\epsilon$-$\gamma$-$\lambda$ parameters). Thus, adapting oneself based on incomplete and possibly partially accurate information is a huge challenge.

\bibliographystyle{ieeetr}

\end{document}